# Calibration of 3-d.o.f. Translational Parallel Manipulators Using Leg Observations

Anatol Pashkevich[1,2], Damien Chablat[1], Philippe Wenger[1], Roman Gomolitsky[2]
[1]*Institut de Recherche en Communications et Cybernétique de Nantes*
*France*
[2]*Belarusian State University of Informatics and Radioelectronics*
*Belarus*

## 1. Introduction

Parallel kinematic machines (PKM) are commonly claimed as appealing solutions in many industrial applications due to their inherent structural rigidity, good payload-to-weight ratio, high dynamic capacities and high accuracy (Tlusty et al., 1999; Tsai, 1999; Merlet, 2000; Wenger et al., 2001). However, while PKM usually exhibit a much better repeatability compared to serial mechanisms, they may not necessarily possess a better accuracy that is limited by manufacturing/assembling errors in numerous links and passive joints (Wang & Masory, 1993). Thus, the PKM accuracy highly relies on an accurate kinematic model, which must be carefully tuned (calibrated) for each manipulator individually.

Similar to serial manipulators, PKM calibration techniques are based on the minimization of a parameter-dependent error function, which incorporates residuals of the kinematic equations (Schröer et al., 1995; Wampler et al., 1995; Fassi et al., 2007; Legnani et al., 2007). For parallel manipulators, the inverse kinematic equations are considered computationally more efficient, contrary to the direct kinematics, which is usually analytically unsolvable for PKM. But the main difficulty with this technique is the full-pose measurement requirement, which is very hard to implement (Innocenti, 1995; Iurascu & Park, 2003; Daney, 2003; Jeong et al., 2004; Huang et al., 2005). Hence, a number of studies have been directed at using the subset of the pose measurement data, which however creates another problem, the identifiability of the model parameters (Khalil & Besnard, 1999; Daney & Emiris, 2001; Besnard & Khalil, 2001; Rauf et al., 2004, 2006).

Popular approaches in parallel robot calibration deal with one-dimensional pose errors using a double-ball-bar system or other measuring devices as well as imposing mechanical constraints on some elements of the manipulator (Zhuang et al., 1999; Thomas et al., 2003; Daney, 1999). However, in spite of hypothetical simplicity, it is hard to implement in practice since an accurate extra mechanism is required to impose these constraints. Additionally, such methods reduce the workspace size and the identification efficiency.

Another category of calibration methods, the self- or autonomous calibration, is implemented by minimizing the residuals between the computed and measured values of the active and/or redundant joint sensors (Hesselbach et al., 2005). Adding extra sensors at usually unmeasured joints is very attractive from a computational point of view, since it allows getting the data in the whole workspace and potentially reduces impact of the measurement noise. However, only a partial set of the parameters may be identified in this way, since the internal sensing is unable to provide sufficient information for the robot end-effector absolute location (Zhuang, 1997; Williams et al., 2006).

More recently, several hybrid calibration methods were proposed that utilize intrinsic properties of a particular parallel machine allowing extracting the full set of the model parameters (or the most essential of them) from a minimum set of measurements. It worth mentioning an innovative approach developed by Renaud et al. (2004 - 2006) who applied the vision-based measurement technique for the parallel manipulators calibration from the leg observations. In this approach, the source data are extracted from the leg images, without any strict assumptions on the end-effector poses. The only assumption is related to the manipulator architecture (the mechanism is actuated by linear drives located on the base). However, current accuracy of the camera-based measurements is not high enough yet to apply this method in industrial environment.

This chapter summarises the authors' results in the area of parallel robotics (Pashkevich et al., 2005, 2006) and focuses on the calibration of the Orthoglide-type mechanisms, which is also actuated by linear drives located on the manipulator base and admits technique of Renaud et al. (2004, 2005). But, in contrast to the known works, our approach assumes that the leg location is observed for specific manipulator postures, when the tool centre point (TCP) moves along the Cartesian axes. For these postures and for the nominal Orthoglide geometry, the legs are strictly parallel to the corresponding Cartesian planes. So, the deviations of the manipulator geometry influence on the leg parallelism that gives the source data for the parameter identification. The main advantage of this approach is the simplicity of the measuring system that can avoid using computer vision and is composed of standard comparator indicators, which are common in industry.

## 2. Orthoglide Mechanism

### 2.1 Manipulator architecture

The Orthoglide is a three d.o.f. parallel manipulator actuated by linear drives with mutually orthogonal axes. Its kinematic architecture is presented in Fig. 1 and includes three identical parallel chains that will be further referred to as "legs". Each manipulator leg is formally described as PRPaR - chain, where P, R and Pa denote the prismatic, revolute, and parallelogram joints respectively. The output machinery (with a tool mounting flange) is connected to the legs in such a manner that the tool moves in the Cartesian space with fixed orientation (i.e. restricted to translational motions). The Orthoglide workspace has a regular, quasi-cubic shape. The input/output equations are simple and the velocity transmission factors are equal to one along the x, y and z direction at the isotropic configuration, like in a conventional serial PPP machine. The latter is an essential advantage for machining applications (Wenger & Chablat, 2000; Chablat & Wenger, 2003).

Another specific feature of the Orthoglide mechanism, which will be further used for calibration, is displayed during the end-effector motions along the Cartesian axes. For example, for the x-axis motion, the sides of the x-leg parallelogram must retain strictly parallel to the x-axis. Hence, the observed deviation may be a data source for calibration.

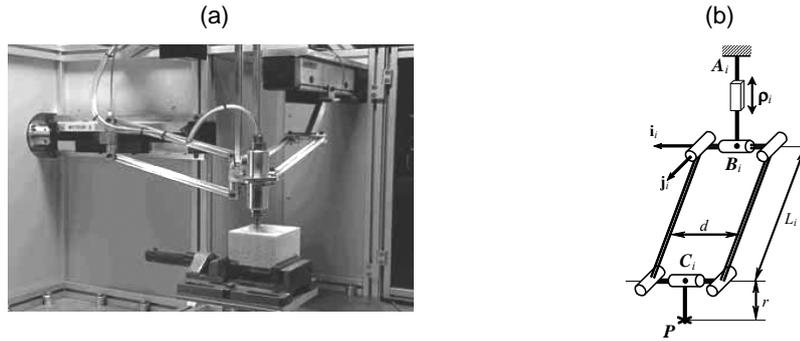

Fig. 1. The architecture of Orthoglide manipulator (a) and kinematics of its leg (b)
(© CNRS Photothèque / CARLSON Leif)

For a small-scale Orthoglide prototype used for the calibration experiments, the workspace size is approximately equal to 200×200×200 mm³ with the velocity transmission factors bounded between 1/2 and 2 (Chablat & Wenger, 2003). The legs nominal geometry is defined by the following parameters: $L$ = 310.25 mm, $d$ = 80 mm, $r$ = 31 mm where $L$, $d$ are the parallelogram length and width, and $r$ is the distance between the points $C_i$ and the tool centre point $P$ (see Fig. 1b).

### 2.2 Modelling assumptions
Following previous studies on the PKM accuracy (Wang & Massory, 1993; Renaud et al., 2004, Caro et al., 2006), the influence of the joint defects is assumed negligible compared to the encoder offsets and the link length deviations. This validates the following modelling assumptions:
(i)   the manipulator parts are supposed to be rigid bodies connected by perfect joints;
(ii)  the manipulator legs (composed of a prismatic joint, a parallelogram, and two revolute joints) generate a four degrees-of-freedom motions;
(iii) the articulated parallelograms are assumed to be perfect but non-identical;
(iv)  the linear actuator axes are mutually orthogonal and are intersected in a single point to ensure a translational movement of the end-effector;
(v)   the actuator encoders are perfect but located with some errors (offsets).

Using these assumptions, calibration equations will be derived based on the observation of the parallel motions of the manipulator legs.

### 2.3 Kinematic model
Since the kinematic parallelograms are admitted to be non-identical, the kinematic model developed in our previous works (Pashkevich et al., 2005, 2006) should be extended to describe the manipulator with different length leg parameters.
Under the adopted assumptions, similar to the equal-leg case, the articulated parallelograms may be replaced by the kinematically equivalent bar links. Besides, a simple transformation of the Cartesian coordinates (shifted by the vector $(r, r, r)^T$, see Fig. 1b) allows us to eliminate the tool offset. Hence, the Orthoglide geometry can be described by a simplified model, which consists of three rigid links connected by spherical joints to the tool centre point at one side and to the allied prismatic joints at another side (Fig. 2). Corresponding formal definition of each leg can be presented as PSS, where P and S denote the actuated prismatic joint and the passive spherical joint respectively.

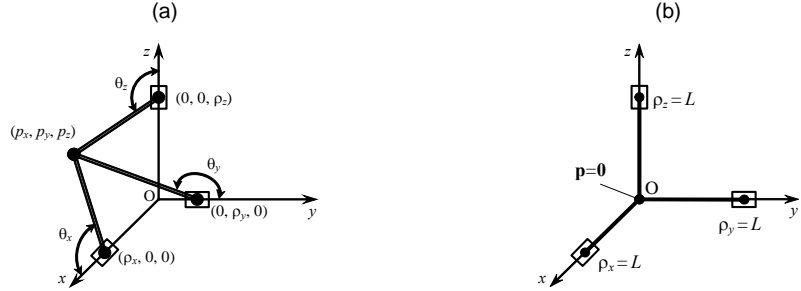

Fig. 2. Orthoglide simplified model (a) and its ''zero'' configuration (b)

Thus, if the origin of a reference frame is located at the intersection of the prismatic joint axes and the $x$, $y$, $z$-axes are directed along them (see Fig. 2), the manipulator kinematics may be described by the following equations

$$\mathbf{p} = \begin{bmatrix} (\rho_x + \Delta\rho_x) + \cos\theta_x \cos\beta_x L_x + e \\ \sin\theta_x \cos\beta_x L_x \\ -\sin\beta_x L_x \end{bmatrix} \quad (1a)$$

$$\mathbf{p} = \begin{bmatrix} -\sin\beta_y L_y \\ (\rho_y + \Delta\rho_y) + \cos\theta_y \cos\beta_y L_y + e \\ \sin\theta_y \cos\beta_y L_y \end{bmatrix} \quad (1b)$$

$$\mathbf{p} = \begin{bmatrix} \sin\theta_z \cos\beta_z L_z \\ -\sin\beta_z L_z \\ (\rho_z + \Delta\rho_z) + \cos\theta_z \cos\beta_z L_z + e \end{bmatrix} \quad (1c)$$

where $\mathbf{p} = (p_x, p_y, p_z)^T$ is the output vector of the TCP position, $\boldsymbol{\rho} = (\rho_x, \rho_y, \rho_z)^T$ is the input vector of the prismatic joints variables, $\Delta\boldsymbol{\rho} = (\Delta\rho_x, \Delta\rho_y, \Delta\rho_z)^T$ is the encoder offset vector, $\theta_i$, $\beta_i$, $i \in \{x, y, z\}$ are the parallelogram orientation angles (internal variables), and $L_i$ are the length of the corresponding leg.

After elimination of the internal variables $\theta_i$, $\beta_i$, the kinematic model (1) can be reduced to three equations

$$\left(p_i - (\rho_i + \Delta\rho_i)\right)^2 + p_j^2 + p_k^2 = L_i^2 \quad (2)$$

which includes components of the input and output vectors **p** and **ρ** only. Here, the subscripts $i, j, k \in \{x, y, z\}$, $i \neq j \neq k$ are used in all combinations, and the joint variables $\rho_i$ are obeyed the prescribed limits $\rho_{min} < \rho_i < \rho_{max}$ defined in the control software (for the Orthoglide prototype, $\rho_{min}$ = -100 mm and $\rho_{max}$ = +60 mm).

It should be noted that, for the case $\Delta\rho_x = \Delta\rho_y = \Delta\rho_z = 0$ and $L_x = L_y = L_z = L$, the nominal "mechanical-zero" posture of the manipulator corresponds to the Cartesian coordinates **p**$_0$ = (0, 0, 0)$^T$ and to the joints variables **ρ**$_0$ = (L, L, L). Moreover, in this posture, the *x*-, y-and *z*-legs are oriented strictly parallel to the corresponding Cartesian axes. But the joint offsets and the leg length differences cause the deviation of the "zero" TCP location and corresponding deviation of the leg parallelism, which may be measured and used for the calibration. Hence, six parameters ($\Delta\rho_x$, $\Delta\rho_y$, $\Delta\rho_z$, $L_x$, $L_y$, $L_z$) define the manipulator geometry and are in the focus of the proposed calibration technique.

### 2.4 Inverse and direct kinematics

The inverse kinematic relations are derived from the equations (2) in a straightforward way and only slightly differ from the "nominal" case:

$$\rho_i = p_i + s_i \sqrt{L_i^2 - p_j^2 - p_k^2} - \Delta\rho_i \qquad (3)$$

where sx, sy, sz $\in \{\pm 1\}$ are the configuration indices defined for the "nominal" geometry as the signs of ρx – px, ρy – py, ρz – pz, respectively. It is obvious that expressions (3) give eight different solutions, however the Orthoglide prototype assembling mode and the joint limits reduce this set to a single case corresponding to sx = sy = sz = 1.

For the direct kinematics, equations (2) can be subtracted pair-to-pair that gives linear relations between the unknowns px, py, pz, which may be expressed in the parametric form as

$$p_i = \frac{\rho_i + \Delta\rho_i}{2} + \frac{t}{\rho_i + \Delta\rho_i} - \frac{L_i^2}{2(\rho_i + \Delta\rho_i)}, \qquad (4)$$

where *t* is an auxiliary scalar variable. This reduces the direct kinematics to the solution of a quadratic equation $At^2 + Bt + C = 0$ with the coefficients

$$A = \sum_{i \neq j}(\rho_i + \Delta\rho_i)^2(\rho_j + \Delta\rho_j)^2; \qquad B = \prod_i(\rho_i + \Delta\rho_i)^2 - \sum_{i \neq j \neq k} L_i^2(\rho_j + \Delta\rho_j)^2(\rho_k + \Delta\rho_k)^2;$$

$$C = \prod_i(\rho_i + \Delta\rho_i)^2 \cdot \left(\sum_i(\rho_i + \Delta\rho_i)^2/4 - \sum_i L_i^2/2\right) + \sum_{i \neq j \neq k} L_i^4(\rho_j + \Delta\rho_j)^2(\rho_k + \Delta\rho_k)^2/4$$

where $i, j, k \in \{x, y, z\}$. From two possible solutions that give the quadratic formula, the Orthoglide prototype (see Fig. 1) admits a single one $t = (-B + \sqrt{B^2 - 4AC})/2A$ corresponding to the selected manipulator assembling mode.

**2.4 Differential relations**

To obtain the calibration equations, let us derive first the differential relations for the TCP deviation for three types of the Orthoglide postures:

(i) "*maximum displacement*" postures for the directions x, y, z (Fig. 3a);

(ii) "*mechanical zero*" or the isotropic posture (Fig. 3b);

(iii) "*minimum displacement*" postures for the directions x, y, z (Fig. 3c);

These postures are of particular interest for the calibration since, in the "nominal" case, a corresponding leg is parallel to the relevant pair of the Cartesian planes.

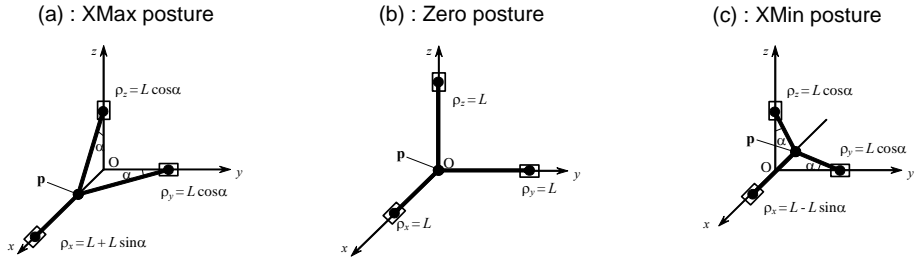

Fig. 3. Specific postures of the Orthoglide (for the x-leg motion along the Cartesian axis X)

The manipulator Jacobian with respect to the parameters $\Delta\boldsymbol{\rho} = (\Delta\rho_x, \Delta\rho_y, \Delta\rho_z)$ and $\mathbf{L} = (L_x, L_y, L_z)$ can be derived by straightforward differentiating of the kinematic equations (2), which yields

$$\begin{bmatrix} p_x - \rho_x & p_y & p_z \\ p_x & p_y - \rho_y & p_z \\ p_x & p_y & p_z - \rho_z \end{bmatrix} \cdot \frac{\partial \mathbf{p}}{\partial \boldsymbol{\rho}} = \begin{bmatrix} p_x - \rho_x & 0 & 0 \\ 0 & p_y - \rho_y & 0 \\ 0 & 0 & p_z - \rho_z \end{bmatrix} ; \quad \begin{bmatrix} p_x - \rho_x & p_y & p_z \\ p_x & p_y - \rho_y & p_z \\ p_x & p_y & p_z - \rho_z \end{bmatrix} \cdot \frac{\partial \mathbf{p}}{\partial \mathbf{L}} = \begin{bmatrix} L_x & 0 & 0 \\ 0 & L_y & 0 \\ 0 & 0 & L_z \end{bmatrix}.$$

Thus, after the matrix inversions and multiplications, the desired Jacobian can be written as

$$\mathbf{J}(\mathbf{p}, \boldsymbol{\rho}) = [\mathbf{J}_\rho(\mathbf{p}, \boldsymbol{\rho}); \ \mathbf{J}_L(\mathbf{p}, \boldsymbol{\rho})] \tag{5}$$

where

$$\mathbf{J}_\rho(.) = \begin{bmatrix} 1 & \dfrac{p_y}{p_x - \rho_x} & \dfrac{p_z}{p_x - \rho_x} \\ \dfrac{p_x}{p_y - \rho_y} & 1 & \dfrac{p_z}{p_y - \rho_y} \\ \dfrac{p_x}{p_z - \rho_z} & \dfrac{p_y}{p_z - \rho_z} & 1 \end{bmatrix}^{-1} \quad \mathbf{J}_L(.) = \begin{bmatrix} \dfrac{p_x - \rho_x}{L_x} & \dfrac{p_y}{L_x} & \dfrac{p_z}{L_x} \\ \dfrac{p_x}{L_y} & \dfrac{p_y - \rho_y}{L_y} & \dfrac{p_z}{L_y} \\ \dfrac{p_x}{L_z} & \dfrac{p_y}{L_z} & \dfrac{p_z - \rho_z}{L_z} \end{bmatrix}^{-1}$$

It should be noted that, for the sake of computing convenience, the above expression includes both the Cartesian coordinates $p_x$, $p_y$, $p_z$ and the joint coordinates $\rho_x$, $\rho_y$, $\rho_z$, but only one of these sets may be treated as independent taking into account the kinematic equations. For the "*Zero*" posture, the differential relations are derived in the neighbourhood of the point {$p_0 = (0, 0, 0)$ ; $\rho_0 = (L, L, L)$}, which after substitution to (5) gives the Jacobian matrix

$$\mathbf{J}_0 = \begin{bmatrix} 1 & 0 & 0 & | & -1 & 0 & 0 \\ 0 & 1 & 0 & | & 0 & -1 & 0 \\ 0 & 0 & 1 & | & 0 & 0 & -1 \end{bmatrix} \qquad (6)$$

Hence, in this case, the TCP displacement is related to the joint offsets and the leg lengths variations $\Delta L_i$ by trivial equations

$$\Delta p_i = \Delta \rho_i - \Delta L_i; \quad i \in \{x, y, z\}. \qquad (7)$$

For the "*XMax*" posture, the Jacobian is computed in the neighbourhood of the point { $\mathbf{p} = (LS_\alpha, 0, 0)$ ; $\boldsymbol{\rho} = (L+LS_\alpha, LC_\alpha, LC_\alpha)$ }, where α is the angle between the *y*-, *z*-legs and the X-axes: $\alpha = a\sin(\rho_{max}/L)$; $S_\alpha = \sin(\alpha)$, $C_\alpha = \cos(\alpha)$. This gives the Jacobian

$$\mathbf{J}_x^+ = \begin{bmatrix} 1 & 0 & 0 & | & -1 & 0 & 0 \\ T_\alpha & 1 & 0 & | & -T_\alpha & -C_\alpha^{-1} & 0 \\ T_\alpha & 0 & 1 & | & -T_\alpha & 0 & -C_\alpha^{-1} \end{bmatrix} \qquad (8)$$

where $T_\alpha = \tan(\alpha)$. Hence, the differential equations for the TCP displacement may be written as $\Delta p_x = \Delta \rho_x - \Delta L_x$

$$\begin{aligned} \Delta p_y &= T_\alpha \Delta \rho_x + \Delta \rho_y - T_\alpha \Delta L_x - C_\alpha^{-1} \Delta L_y \\ \Delta p_z &= T_\alpha \Delta \rho_x + \Delta \rho_z - T_\alpha \Delta L_x - C_\alpha^{-1} \Delta L_z \end{aligned} \qquad (9)$$

It can be proved that similar results are valid for the *YMax* and *ZMax* postures (differing by the indices only), and also for the *XMin*, *YMin*, *ZMin* postures. In the latter case, the angle α should be computed as $\alpha = \operatorname{asin}(\rho_{min}/L)$.

## 3. Calibration Method

### 3.1 Measurement technique
To identify the Orthoglide kinematic parameters specified in the previous section, two approaches can be used, which employ different measurement techniques to evaluate the leg-to-surface parallelism. The first of them (Fig. 4a) assumes two measurements for the same leg posture (to assess distances from both leg ends to the base surface). The second technique (Fig. 4b) assumes a fixed location of the measuring device but two distinct leg postures, which are assumed to be parallel to each other in the nominal case.
It is obvious that, for the perfectly calibrated manipulator, both methods give zero differences for each measurement pair. In contrasts, the non-zero differences contain source information for the parameter identification. However, the first method involves absolute measurements that require essential implementation efforts; besides it allows evaluating parallelism only for the X- and Y-legs with respect to the XY-plane. So, the second method will be used here.

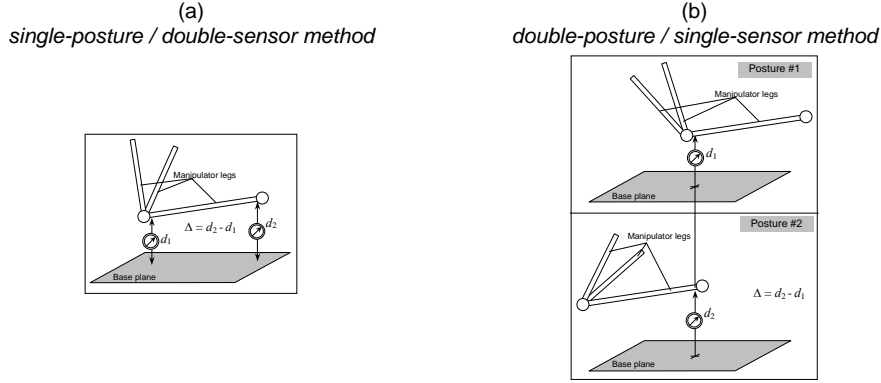

(a) *single-posture / double-sensor method*   (b) *double-posture / single-sensor method*

Fig. 4. Measuring the leg parallelism with respect to the base plane

For this method, which employs the relative measurements and allows assessing the leg parallelism with respect to both relevant planes (XY- and XZ-planes for the X-leg, for instance), the calibration experiment may be arranged in the following way:

**Step 1**. Move the manipulator to the *Zero* posture; locate two gauges in the middle of the X-leg (orthogonal to the leg and parallel to the axes Y and Z); get their readings.

**Step 2**. Move the manipulator sequentially to the *XMax* and *XMin* postures, get the gauge readings, and compute the differences $\Delta y_x^+$, $\Delta z_x^+$, $\Delta y_x^-$, $\Delta z_x^-$ with respect to the "Zero" posture values.

**Step 3+**. Repeat steps 1, 2 for the Y- and Z-legs and compute the differences $\Delta x_y^+$, $\Delta z_y^+$, $\Delta x_y^-$, $\Delta z_y^-$, and $\Delta x_z^+$, $\Delta y_z^+$, $\Delta x_z^-$, $\Delta y_z^-$ corresponding to these legs.

In the above description, the variable following the symbol Δ denotes the measurement direction (x, y or z), the subscript defines the manipulator leg, and the superscript indicates the manipulator posture ('+' for *XMax* and '-' for *XMin*). For example, $\Delta z_x^+$ denotes the z-coordinate deviation of the X-leg for the *XMax* posture with respect to *Zero* location.

### 3.2 Calibration equations

The system of calibration equations can be derived in two steps. First, it is required to define the gauges' initial locations that are assumed to be positioned at the leg middle at the *Zero* posture, i.e. at the points $(\mathbf{p}+\mathbf{r}_i)/2$, $i \in \{x, y, z\}$ where the vectors $r_i$ define the prismatic joints centres: $\mathbf{r}_x = (L+\Delta\rho_x; 0; 0)^T$; $\mathbf{r}_y = (0; L+\Delta\rho_y; 0)^T$; $\mathbf{r}_z = (0; 0; L+\Delta\rho_z)^T$. Hence, using the equation (7), the gauge initial locations can be expressed as

$$\mathbf{g}_x^0 = \left[ (L-\Delta L_x)/2 + \Delta\rho_x; \quad (\Delta\rho_y - \Delta L_y)/2; \quad (\Delta\rho_z - \Delta L_z)/2 \right]^T$$

$$\mathbf{g}_y^0 = \left[ (\Delta\rho_x - \Delta L_x)/2; \quad (L - \Delta L_y)/2 + \Delta\rho_y; \quad (\Delta\rho_z - \Delta L_z)/2 \right]^T \tag{10}$$

$$\mathbf{g}_z^0 = \left[ (\Delta\rho_x - \Delta L_x)/2; \quad (\Delta\rho_y - \Delta L_y)/2; \quad (L - \Delta L_z)/2 + \Delta\rho_z \right]^T$$

Afterwards, for the *XMax*, *YMax*, *ZMax* postures, the leg spatial location is also defined by two points, namely, (i) the tool centre point $\mathbf{p}$, and (ii) the centre of the prismatic joint $\mathbf{r}_i$. For example, for the *XMax* posture, the TCP position is $\mathbf{p}_x^{max} = (LS_\alpha + \Delta\rho_x - \Delta L_x; \ *; \ *)$, the prismatic joint position is $\mathbf{r}_x^{max} = (L + LS_\alpha + \Delta\rho_x; \ 0; \ 0)$. So, the leg is located along the line

$$\mathbf{s}_x(\mu) = \mu \cdot \mathbf{p}_x^{max} + (1-\mu) \cdot \mathbf{r}_x^{max}; \quad \mu \in [0;1],$$

Since the x-coordinate of the gauge is kept constant (for X-leg measurements), the parameter $\mu$ may be obtained from the equation $[\mathbf{s}_x(\mu)]_x = [\mathbf{g}_x^0]_x$, which yields:

$$\mu = 0.5 + S_\alpha - S_\alpha \cdot \Delta L_x / L,$$

Hence, after some transformations, the deviations of the *X*-leg measurements (between the *XMax* and *Zero* postures) may be expressed as

$$\Delta y_x^+ = (0.5 + S_\alpha)T_\alpha \Delta\rho_x + S_\alpha \Delta\rho_y - (0.5 + S_\alpha)T_\alpha \Delta L_x - ((0.5 + S_\alpha)C_\alpha^{-1} - 0.5)\Delta L_y$$

$$\Delta z_x^+ = (0.5 + S_\alpha)T_\alpha \Delta\rho_x + S_\alpha \Delta\rho_z - (0.5 + S_\alpha)T_\alpha \Delta L_x - ((0.5 + S_\alpha)C_\alpha^{-1} - 0.5)\Delta L_z$$

A similar approach may be applied to the *XMin* posture, as well as to the corresponding postures for the Y- and Z-legs. This gives the system of twelve linear equations in six unknowns:

$$\begin{bmatrix} a_1 & b_1 & 0 & -c_1 & -b_1 & 0 \\ b_1 & a_1 & 0 & -b_1 & -c_1 & 0 \\ a_2 & b_2 & 0 & -c_2 & -b_2 & 0 \\ b_2 & a_2 & 0 & -b_2 & -c_2 & 0 \\ 0 & a_1 & b_1 & 0 & -c_1 & -b_1 \\ 0 & b_1 & a_1 & 0 & -b_1 & -c_1 \\ 0 & a_2 & b_2 & 0 & -c_2 & -b_2 \\ 0 & b_2 & a_2 & 0 & -b_2 & -c_2 \\ a_1 & 0 & b_1 & -c_1 & 0 & -b_1 \\ b_1 & 0 & a_1 & -b_1 & 0 & -c_1 \\ a_2 & 0 & b_2 & -c_2 & 0 & -b_2 \\ b_2 & 0 & a_2 & -b_2 & 0 & -c_2 \end{bmatrix} \cdot \begin{bmatrix} \Delta\rho_x \\ \Delta\rho_y \\ \Delta\rho_z \\ \Delta L_x \\ \Delta L_y \\ \Delta L_z \end{bmatrix} = \begin{bmatrix} \Delta x_y^+ \\ \Delta y_x^+ \\ \Delta x_y^- \\ \Delta y_x^- \\ \Delta y_z^+ \\ \Delta z_y^+ \\ \Delta y_z^- \\ \Delta z_y^- \\ \Delta x_z^+ \\ \Delta z_x^+ \\ \Delta x_z^- \\ \Delta z_x^- \end{bmatrix} \tag{11}$$

where $a_i = \sin(\alpha_i)$;  $b_i = (0.5 + \sin(\alpha_i))\tan(\alpha_i)$;  $c_i = (0.5 + \sin(\alpha_i))/\cos(\alpha_i) - 0.5$;  $i \in \{1,2\}$, and $\alpha_1 = \mathrm{asin}(\rho_{max}/L) > 0$;  $\alpha_2 = \mathrm{asin}(\rho_{min}/L) < 0$. This system can be solved using the pseudoinverse of Moore-Penrose, which ensures the minimum of the residual square sum for corresponding linear approximation of the kinematic equations that is valid for small values of $\Delta\rho_x, \Delta\rho_y, \ldots \Delta L_y, \Delta L_z$. Otherwise, it is prudent to apply straightforward numerical optimisation, which fits the experimental data to the manipulator kinematic model (1).

**3.4 Calibration accuracy**

Because of the measurement noise, the developed technique may produce some errors in estimates of the model parameters. Thus, for practical applications, it is worth to evaluate the statistical properties of the calibration errors.

Within the linear calibration equations (11), the impact of the measurement noise may be evaluated using general techniques from the identification theory, under the standard assumptions concerning the primary measurement errors $\xi(.)$ (zero-mean independent and identically distributed Gaussian random variables with the standard deviation σ). For these assumptions, the covariance matrix of the estimated parameters is written as (Ljung, 1999)

$$\mathbf{V}(\Delta\boldsymbol{\rho}, \Delta\mathbf{L}) = (\mathbf{J}^T\mathbf{J})^{-1} \cdot \mathbf{J}^T \cdot \mathbf{E}(\Delta\mathbf{s} \cdot \Delta\mathbf{s}^T) \cdot \mathbf{J} \cdot (\mathbf{J}^T\mathbf{J})^{-1} \qquad (12)$$

where $\mathbf{E}(.)$ denotes the mathematical expectation, $\mathbf{J}$ is the identification Jacobian, and $\Delta\mathbf{s}$ is the vector of the measurement errors in the right-hand side of the system (11). However, in contrast to the standard technique, the vector $\Delta\mathbf{s}$ includes some statistically-dependent components because the same measurement values, corresponding to the Zero position, are subtracted from those corresponding to the Max and Min postures. In particular,

$$\Delta\mathbf{s} = \left[\xi(x_y^+) - \xi(x_y^0),\ \xi(y_x^+) - \xi(y_x^0), \ldots\ \xi(z_x^+) - \xi(z_x^0)\right]^T, \qquad (13)$$

where the index sequence strictly corresponds to (11). Thus, the covariance $\mathbf{E}(\Delta\mathbf{s} \cdot \Delta\mathbf{s}^T)$ is the 12×12 non-identity matrix that after relevant transformations may be expressed as

$$\mathbf{E}\left(\Delta\mathbf{s} \cdot \Delta\mathbf{s}^T\right) = \sigma^2 \cdot \begin{bmatrix} \mathbf{G} & \mathbf{0} & \mathbf{0} \\ \mathbf{0} & \mathbf{G} & \mathbf{0} \\ \mathbf{0} & \mathbf{0} & \mathbf{G} \end{bmatrix}_{12\times 12}; \qquad \mathbf{G} = \begin{bmatrix} 2 & 0 & 1 & 0 \\ 0 & 2 & 0 & 1 \\ 1 & 0 & 2 & 0 \\ 0 & 1 & 0 & 2 \end{bmatrix} \qquad (14)$$

Hence, using expressions (12), (14) it is possible to evaluate the identification accuracy (via the covariance matrix (12)) for the set of parameters $\{\Delta\rho_x, \Delta\rho_y, \ldots \Delta L_y, \Delta L_z\}$ provided the measurement error parameter σ is known. For instance, for the Orthoglide prototype described in sub-section 2.1 and the Max/Min posture characteristic angles $\alpha_1 = 11.0°$ and $\alpha_2 = -18.7°$, the measurement noise with $\sigma = 10^{-2}$ mm causes the mean-square errors for the $\Delta\rho_x, \Delta\rho_y, \ldots \Delta L_y, \Delta L_z$ of about 0.07 mm.

## 4. Experimental results

### 4.1 Experimental setup

For experimental verification of the developed technique, we used the measuring system composed of standard comparator indicators with resolution of 10.0 μm. The indicators were attached to universal magnetic stands that allow fixing them on the manipulator base.

This system is sequentially used for measuring the *X-*, *Y-*, and *Z*-leg parallelism while the manipulator moves between the *Max*, *Min* and *Zero* postures. (It is obvious that for industrial applications it is worth using more sophisticated digital indicators with the resolution of 1.0 μm or less, which yield more accurate calibration results.)

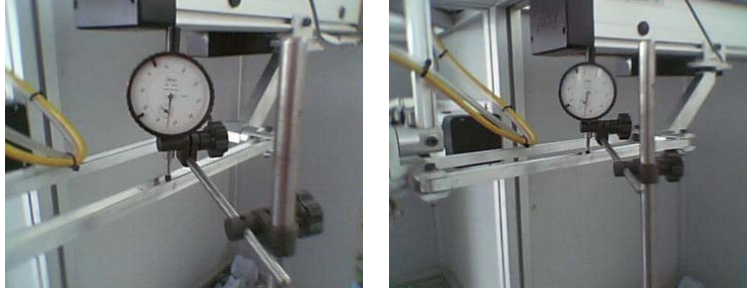

Fig. 5. Experimental setup for calibration experiments

For each measurement, the indicators are located on the mechanism base in such a manner that a corresponding leg is admissible for the gauge contact for all intermediate postures (Fig. 5). The *Min* and *Max* postures are constrained by the software limits and defined as $\rho_{min} = -100.00$ mm and $\rho_{max} = +60.00$ mm respectively. Initial position of the indicator corresponds to the leg middle point at the manipulator *Zero* posture.

During experiments, the legs were moved sequentially via the following postures: Zero → Max → Min → Zero→ … . To reduce the measurement errors, the measurements were repeated three times for each leg. Then, the results were averaged and used for the parameter identification. It should be noted that the measurements demonstrated very high repeatability compared to the encoder resolution (dissimilarity was less than 20.0 μm).

**4.2 Calibration results and their analysis**

The experimental study included three types of experiments targeted to the following objectives: (#1) validation of modelling assumptions; (#2) obtaining source data for the parameter identification; and (#3) verification of the calibration results.

**Experiment #1**. The first calibration experiment demonstrated rather high parallelism deviation for the legs at the *Max* and *Min* postures, up to 2.37 mm as shown in Table1. This indicated low accuracy of the nominal kinematic model and motivated necessity of the calibration. On the other hand, the milling accuracy evaluated in separate tests was quite good. However, this is not an indicator of high absolute accuracy but just a proof of the Orthoglide architecture advantages (the milling tests were perfect just because of the high homogeneity of the manipulator workspace in the neighbourhood of the isotropic location).

The straightforward application of the proposed calibration algorithm to the data set #1 was not optimistic: in the frames of the adopted kinematic model the root-mean-square (r.m.s.) deviation for the legs can be reduced down from 1.19 mm to 0.74 mm only (see Table 1 where $\Delta x_y = \Delta x_y^+ - \Delta x_y^-$, $\Delta x_z = \Delta x_z^+ - \Delta x_z^-$, etc.). Besides, the statistical estimation of the measurement noise parameter σ (based on the residual analysis) also yielded unrealistic result compared to the encoder resolutions (0.01 mm). This impelled to conclude that some modelling assumptions are not valid and the manipulator mechanics required more careful tuning, especially orientation of the linear actuator axes (that are assumed to be mutually orthogonal and to intersect in a single point). Thus, the manipulator mechanics was re-tuned, in particular spatial locations of the actuator axes were adjusted.

| Data Source | $\Delta x_y$ | $\Delta x_z$ | $\Delta y_x$ | $\Delta y_z$ | $\Delta z_x$ | $\Delta z_y$ | r.m.s. |
|---|---|---|---|---|---|---|---|
| *Experiment #1 (before mechanical tuning and before calibration)* | | | | | | | |
| Measurements #1 | +0.52 | +1.58 | +2.37 | -0.25 | -0.57 | -0.04 | 1.19 |
| Expected improvement | -0.94 | +0.63 | +1.07 | -0.84 | -0.27 | +0.35 | 0.74 |
| *Experiment #2 (after mechanical tuning, before calibration)* | | | | | | | |
| Measurements #2 | -0.43 | -0.37 | +0.42 | -0.18 | -1.14 | -0.70 | 0.62 |
| Expected improvement | -0.28 | +0.25 | +0.21 | -0.14 | -0.13 | +0.09 | 0.20 |
| *Experiment #3 (after calibration and adjusting of $\Delta\rho$)* | | | | | | | |
| Measurements #3 | -0.23 | +0.27 | +0.34 | -0.10 | -0.09 | +0.11 | 0.21 |
| Expected improvement | -0.29 | +0.23 | +0.25 | -0.17 | -0.10 | +0.08 | 0.20 |

Table 1. Experimental data and expected improvements of accuracy via calibration [mm]

**Experiment #2.** The second calibration experiment (after mechanical tuning) yielded lower parallelism deviations, less than 0.62 mm in terms of the deviations $\Delta x_y$, $\Delta x_z$, ... (see Table 1), which is about twice better than in the first experiment. Besides, the expected residual reduction was also essential (0.20 mm) that justified validity of the modelling assumptions. For these data, the developed calibration algorithm was applied for three sets of the model parameters: for the full set {$\Delta\rho$, $\Delta L$} and for the reduced sets {$\Delta\rho$}, {$\Delta L$}. As follows from the identification results (Tables 2, 3), the calibration algorithm is able to identify simultaneously both the joint offsets and $\Delta\rho$ and the link lengths $\Delta L$. However, both $\Delta\rho$ and $\Delta L$ (separately) demonstrate roughly the same influence on the residual reduction, from 0.32 mm to 0.14 mm (in terms of the deviations $\Delta x_y^+, \Delta x_y^-, \Delta x_z^+, \Delta x_z^-, \ldots$), while the full set {$\Delta\rho$, $\Delta L$} gives further residual reduction down to 0.12 mm only. This motivates considering $\Delta\rho$ as the most essential parameters to be calibrated. Accordingly, the identified vales of joint offsets $\Delta\rho_x$, $\Delta\rho_y$, $\Delta\rho_z$ were incorporated in the Orthoglide control software.

**Experiment #3.** The third experiment was targeted to the validation of the calibration results, i.e. assessing the leg parallelism while using the kinematic model with the parameters identified from the data set #2. This experiment demonstrated very good agreement with the expected values of $\Delta x_y^+, \Delta x_y^-, \Delta x_z^+, \Delta x_z^-, \ldots$. In particular, the maximum deviation reduced down from 0.62 mm to 0.24 mm, and the root-mean-square value decreased down from 0.32 mm to 0.15 mm (expected value is 0.14 mm). On the other hand, further fitting of the kinematic model to the third data set gives both negligible improvement in the deviations and very small alteration of the model parameters. It is evident that further reduction of the parallelism deviation is bounded by the manufacturing errors and, by non-geometric reasons.

| Residuals | Experiment 2 | | | | Experiment 3 | | | |
|---|---|---|---|---|---|---|---|---|
| | *Exper. data* | *Expected improvement* | | | *Exper. data* | *Expected improvement* | | |
| | | {Δρ, ΔL} | {Δρ} | {ΔL} | | {Δρ, ΔL} | {Δρ} | {ΔL} |
| $\Delta x_y^+$ | −0.19 | -0.09 | -0.03 | -0.03 | -0.07 | 0.02 | 0.04 | 0.04 |
| $\Delta y_x^+$ | 0.08 | 0.12 | 0.03 | 0.04 | 0.02 | 0.04 | 0.02 | 0.02 |
| $\Delta x_y^-$ | 0.22 | 0.09 | 0.13 | 0.12 | 0.10 | -0.07 | -0.06 | -0.06 |
| $\Delta y_x^-$ | -0.34 | -0.10 | -0.13 | -0.13 | -0.24 | 0.01 | 0.00 | 0.00 |
| $\Delta x_z^+$ | -0.29 | -0.41 | -0.32 | -0.33 | 0.01 | -0.02 | -0.01 | 0.00 |
| $\Delta z_x^+$ | -0.52 | -0.45 | -0.39 | 0.42 | 0.11 | -0.02 | -0.03 | 0.04 |
| $\Delta x_z^-$ | 0.08 | 0.23 | 0.26 | 0.26 | -0.19 | -0.05 | -0.04 | -0.04 |
| $\Delta z_x^-$ | 0.62 | 0.55 | 0.57 | 0.56 | -0.03 | 0.10 | 0.09 | 0.09 |
| $\Delta y_z^+$ | 0.02 | -0.04 | -0.13 | -0.12 | 0.07 | -0.03 | -0.05 | -0.05 |
| $\Delta z_y^+$ | -0.24 | 0.29 | -0.26 | -0.27 | -0.21 | -0.05 | -0.07 | -0.07 |
| $\Delta y_z^-$ | 0.20 | -0.03 | 0.06 | -0.06 | 0.17 | 0.04 | 0.03 | 0.03 |
| $\Delta z_y^-$ | 0.45 | 0.48 | 0.51 | 0.50 | 0.27 | 0.11 | 0.10 | 0.10 |
| *Average* | 0.32 | 0.12 | 0.14 | 0.14 | 0.15 | 0.13 | 0.14 | 0.14 |

Table 2. Residual compensation using different sets of kinematic parameters [mm]

| Set of parameters | Identified values [mm] | | | | | | Residuals |
|---|---|---|---|---|---|---|---|
| | $\Delta\rho_x$ | $\Delta\rho_y$ | $\Delta\rho_z$ | $\Delta L_x$ | $\Delta L_y$ | $\Delta L_z$ | |
| {Δρ, ΔL} | 4.66 | -5.36 | 1.46 | 5.20 | -5.96 | 3.16 | 0.12 |
| {Δρ} | -0.48 | 0.49 | -1.67 | – | – | – | 0.14 |
| {ΔL} | – | – | – | 0.50 | -0.52 | 1.69 | 0.14 |

Table3. Calibration results for parameters Δ**ρ** and Δ*L*

**Resume**. Hence, the calibration results confirm validity of the proposed identification technique and its ability to tune the joint offsets and link lengths from observations of the leg parallelism. However, for these partucular experiments, combined influence of the parameters {Δ**ρ**, Δ*L*} may be roughly decribed by the diffrence {Δ**ρ** - Δ*L*} that allows us to simplify modifications of the kinematic model included in the control software. Another conclusion is related to the modelling assumption: for further accuracy improvement it is prudent to generalize the manipulator model by including parameters describing orientation of the prismatic joint axes, which is equavalet to relaxing some modelling assumption.

## 5. Conclusions

Recent advances in parallel robot architectures encourage related research on kinematic calibration of parallel mechanisms. This paper proposes a novel calibration approach based on observations of manipulator leg parallelism with respect to the Cartesian planes. Presented for the Orthoglide-type mechanisms, this approach may be also applied to other manipulator architectures that admit parallel leg motions (along the Cartesian axes) or, in more general case, allow locating the leg in several postures with a common intersection point.

The proposed calibration technique employs a simple and low-cost measuring system composed of standard comparator indicators attached to the universal magnetic stands. They are sequentially used for measuring the deviation of the relevant leg location while the manipulator moves the tool-center-point in the directions x, y and z. From the measured differences, the calibration algorithm estimates the joint offsets and link lengths that are treated as the most essential parameters that are difficult to identify by other methods.

The presented theoretical derivations deal with the sensitivity analysis of the proposed measurement method and also with the calibration accuracy. The validity of the proposed approach and efficiency of the developed numerical algorithm were confirmed by the calibration experiments with the Orthoglide prototype, which allowed dividing the residual root-mean-square by three.

To increase the calibration precision, future work will focus on the development of the specific assembling fixture ensuring proper location of the linear actuators and also on the expanding the set of the identified model parameters and compensation of the non-geometric errors that are not identified within the frames of the adopted model.

## 6. References

Besnard, S. & Khalil, W. (2001). Identifiable parameters for parallel robots kinematic calibration. In: *IEEE International Conference on Robotics and Automation*, Vol. 3, pp. 2859-2866, May 2001, Seoul, Korea.

Caro, S., Wenger, P., Bennis, F. & Chablat, D. (2006). Sensitivity Analysis of the Orthoglide, a 3-DOF Translational Parallel Kinematic Machine. *ASME Journal of Mechanical Design*, Vol. 128, No 2, (March 2006), 392-402.

Chablat, D. & Wenger, P. (2003). Architecture Optimization of a 3-DOF Parallel Mechanism for Machining Applications, the Orthoglide. *IEEE Transactions on Robotics and Automation*, Vol. 19, No 3, (June 2003), 403-410.

Daney, D. (1999). Self calibration of Gough platform using leg mobility constraints. In: *Proceedings of the 10th World Congress on the Theory of Machine and Mechanisms*, pp. 104–109, June 1999, Oulu, Finland.

Daney, D. (2003). Kinematic Calibration of the Gough platform. *Robotica*, Vol. 21, No 6, (Dec. 2003), 677-690.

Daney, D. & Emiris I.Z. (2001). Robust parallel robot calibration with partial information. In: *IEEE International Conference on Robotics and Automation*, Vol. 4, pp. 3262-3267, May 2001, Seoul, Korea.

Fassi I., Legnani G., Tosi D. & Omodei A. (2007). Calibration of Serial Manipulators: Theory and Applications. In: *Industrial Robotics: Programming, Simulation and Applications*, Proliteratur Verlag, Mammendorf, Germany, pp. 147 - 170.

Hesselbach, J., Bier, C., Pietsch, I., Plitea, N., Büttgenbach, S., Wogersien, A. & Güttler, J. (2005). Passive-joint sensors for parallel robots. *Mechatronics*, Vol. 15, No 1, (Feb. 2005), 43-65.

Huang, T., Chetwynd, D. G., Whitehouse, D. J., & Wang, J. (2005). A general and novel approach for parameter identification of 6-dof parallel kinematic machines. *Mechanism and Machine Theory*, Vol. 40, No 2, (Feb. 2005), 219-239.

Innocenti, C. (1995). Algorithms for kinematic calibration of fully-parallel manipulators In: *Computational Kinematics*, J-P. Merlet and B. Ravani (Eds.), pp. 241-250, Dordrecht: Kluwer Academic Publishers.

Iurascu, C.C. & Park, F.C. (2003). Geometric algorithm for kinematic calibration of robots containing closed loops. *ASME Journal of Mechanical Design*, Vol. 125, No 1, (March 2003), 23-32.

Jeong, J., Kang, D., Cho, Y.M., & Kim, J. (2004). Kinematic calibration of redundantly actuated parallel mechanisms. *ASME Journal of Mechanical Design*, Vol. 126, No 2, (March 2004), 307-318.

Khalil, W. & Besnard, S. (1999). Self calibration of Stewart–Gough parallel robots without extra sensors. *IEEE Transactions on Robotics and Automation*, Vol. 15, No 6, (Dec. 1999), 1116–1112.

Legnani, G., Tosi; D., Adamini, R. & Fassi, I.. (2007). Calibration of Parallel Kinematic Machines: theory and applications. In: *Industrial Robotics: Programming, Simulation and Applications*, Proliteratur Verlag, Mammendorf, Germany, pp. 171 - 194.

Ljung, L. (1999). *System identification : theory for the user* (2nd ed). Prentice Hall, New Jersey.

Merlet, J.-P. (2000). *Parallel Robots*. Kluwer Academic Publishers, Dordrecht.

Pashkevich, A., Wenger, P. & Chablat, D. (2005). Design strategies for the geometric synthesis of Orthoglide-type mechanisms. *Mechanism and Machine Theory*, Vol. 40, No 8, (Aug. 2005), 907-930.

Pashkevich, A., Chablat, D. & Wenger, P. (2006). Kinematics and workspace analysis of a three-axis parallel manipulator: the Orthoglide. *Robotica*, Vol. 4, No 1, (Jan. 2006), 39-49.

Pashkevich, A., Chablat, D. & Wenger, P. (2006) Kinematic Calibration of Orthoglide-Type Mechanisms. Information Control Problems in Manufacturing 2006 (A Proceedings Volume from the 12th IFAC Conference 17-19 May 2006, Saint-Etienne, France), pp. 149-154, 2006

Rauf, A., Kim, S.-G. & Ryu, J. (2004). Complete parameter identification of parallel manipulators with partial pose information using a new measurements device. *Robotica*, Vol. 22, No 6, (Nov. 2004), 689-695.

Rauf, A., Pervez A. & Ryu, J. (2006). Experimental results on kinematic calibration of parallel manipulators using a partial pose measurement device. *IEEE Transactions on Robotics*, Vol. 22, No 2, (Apr. 2006), 379-384.

Renaud, P., Andreff, N., Pierrot, F., & Martinet, P. (2004). Combining end-effector and legs observation for kinematic calibration of parallel mechanisms. In: *IEEE International Conference on Robotics and Automation*, Vol. 4, pp. 4116-4121, Apr.-May 2004, New Orleans, USA.

Renaud, P., Andreff, N., Gogu, G. & Martinet, P. (2005). Kinematic calibration of parallel mechanisms: a novel approach using legs observation. *IEEE Transactions on Robotics*, Vol. 21, No 4, (Aug. 2005), 529-538.

Renaud, P., Vivas, A., Andreff, N., Poignet, P., Martinet, P., Pierrot, F. & Company, O. (2006). Kinematic and dynamic identification of parallel mechanisms. *Control Engineering Practice*, Vol. 14, No 9, (Sept. 2006), 1099-1109

Schröer, K., Bernhardt, R., Albright, S., Wörn, H., Kyle, S., van Albada, D., Smyth, J. & Meyer, R. (1995). Calibration applied to quality control in robot production. *Control Engineering Practice*, Vol. 3; No 4, (Apr. 1995), 575-580.

Thomas, F., Ottaviano, E., Ros, L., & Ceccarelli, M. (2005). Performance analysis of a 3–2–1 pose estimation device. *IEEE Transactions on Robotics*, Vol. 21, No 3, (June 2005), 88-297.

Tlusty, J., Ziegert, J.C. & Ridgeway, S. . (1999). Fundamental comparison of the use of serial and parallel kinematics for machine tools, *CIRP Annals*, Vol. 48, No 1, 351-356.

Tsai, L. W. (1999). *Robot analysis: the mechanics of serial and parallel manipulators*. John Wiley & Sons, New York.

Wampler, C.W., Hollerbach, T.M. & Arai, T. (1995). An implicit loop method for kinematic calibration and its application to closed chain mechanisms. *IEEE Transactions on Robotics and Automation*, Vol. 11, No 5, (Oct. 1995), 710–724.

Wang, J. & Masory, O. (1993). On the accuracy of a Stewart platform - Part I: The effect of manufacturing tolerances. In: *IEEE International Conference on Robotics and Automation*, Vol. 1, pp. 114–120, May 1993, Atlanta, USA.

Wenger, P. & Chablat, D. (2000). Kinematic analysis of a new parallel machine-tool: the Orthoglide. In: *7th International Symposium on Advances in Robot Kinematics*, pp. 305-314, June 2000, Portoroz, Slovenie.

Wenger, P., Gosselin, C. & Chablat, D. (2001). Comparative study of parallel kinematic architectures for machining applications. In: *Workshop on Computational Kinematics*, pp. 249-258, May 2001, Seoul, Korea.

Williams, I., Hovland, G. & Brogardh, T. (2006). Kinematic error calibration of the gantry-tau parallel manipulator. In: *IEEE International Conference on Robotics and Automation*, pp. 4199-4204, May 2006, Orlando, USA.

Zhuang, H. (1997). Self-calibration of parallel mechanisms with a case study on Stewart platforms. *IEEE Transactions on Robotics and Automation*, Vol. 13, No 3, (June 1997), 387–397.

Zhuang, H., Motaghedi, S.H. & Roth, Z.S. (1999). Robot calibration with planar constraints. In: *IEEE International Conference of Robotics and Automation*, Vol. 1, pp. 805-810, May 1999, Detroit, USA.